\documentclass[pdflatex,sn-mathphys-num]{sn-jnl}

\usepackage{graphicx}%
\usepackage{multirow}%
\usepackage{amsmath,amssymb,amsfonts}%
\usepackage{amsthm}%
\usepackage{mathrsfs}%
\usepackage[title]{appendix}%
\usepackage{xcolor}%
\usepackage{textcomp}%
\usepackage{manyfoot}%
\usepackage{booktabs}%
\usepackage{algorithm}%
\usepackage{algorithmicx}%
\usepackage{algpseudocode}%
\usepackage{listings}%
\usepackage{bbm}

\theoremstyle{thmstyleone}%

\theoremstyle{thmstyletwo}%

\theoremstyle{thmstylethree}%

\begin{document}

\title[Article Title]{Who Does This Name Remind You of ? Nationality Prediction via Large Language Model Associative Memory}

\author*[1,2]{\fnm{Keito} \sur{Inoshita}}\email{inosita.2865@gmail.com}

\affil*[1]{\orgdiv{Faculty of Business and Commerce}, \orgname{Kansai University}, \orgaddress{\street{3-3-35 Yamatecho}, \city{Suita}, \postcode{5648680}, \state{Osaka}, \country{Japan}}}

\affil[2]{\orgdiv{Data Science and AI Innovation Research Promotion Center}, \orgname{Shiga University}, \orgaddress{\street{1-1-1 Baba}, \city{Hikone}, \postcode{5228522}, \state{Shiga}, \country{Japan}}}

\abstract{Large language models (LLMs) possess extensive world knowledge, yet methods for effectively eliciting this knowledge remain underexplored. Nationality and region prediction tasks require understanding of not only linguistic features but also cultural and historical background, making LLM world knowledge particularly valuable. However, conventional LLM prompting methods rely on direct reasoning approaches, which have limitations in applying abstract linguistic rules. We propose LLM Associative Memory Agents (LAMA), a novel framework that leverages LLM world knowledge as associative memory. Rather than directly inferring nationality from names, LAMA recalls famous individuals with the same name and aggregates their nationalities through indirect reasoning. A dual-agent architecture comprising a Person Agent and a Media Agent, specialized in different knowledge domains, recalls famous individuals in parallel, generating Top-1 predictions through voting and Top-K predictions through conditional completion. On a 99-country nationality prediction task, LAMA achieved 0.817 accuracy, substantially outperforming conventional LLM prompting methods and neural models. Our experiments reveal that LLMs exhibit higher reliability in recalling concrete examples than in abstract reasoning, that recall-based approaches are robust to low-frequency nationalities independent of data frequency distributions, and that the dual-agent architecture functions complementarily to produce synergistic effects. These results demonstrate the effectiveness of a new multi-agent system that retrieves and aggregates LLM knowledge rather than prompting reasoning.
}


\maketitle

\section{Introduction}

The emergence of large language models (LLMs) has triggered a paradigm shift across numerous natural language processing tasks. State-of-the-art LLMs such as GPT-4, Claude, and Gemini are pre-trained on text corpora spanning trillions of tokens, acquiring extensive world knowledge encompassing language, culture, geography, and history \cite{1}. How to effectively elicit and leverage this world knowledge represents one of the central challenges in LLM research. Conventional approaches have predominantly relied on directly querying LLMs through prompts and depending on their reasoning capabilities to obtain answers. Prompting techniques such as Chain-of-Thought (CoT) \cite{2} and Self-Reflection \cite{3} have achieved performance improvements by making LLM reasoning processes explicit and refining them. However, these methods remain optimizations within the framework of having LLMs reason directly, rather than fundamentally reconsidering how LLM knowledge should be utilized.

Nationality and region prediction is the task of estimating an individual's nationality or region of origin from their personal name. This task has broad practical applications, including customer analysis in marketing \cite{4}, ethnic composition estimation in demographic research \cite{5}, and ancestor tracing in genealogical studies. Conventionally, two approaches have been employed for this task. The first is statistical pattern learning using neural models. Various methods have been proposed, ranging from logistic regression with character n-gram features \cite{6}, to deep learning with RNNs \cite{7}, to fine-tuning pre-trained models such as CANINE \cite{8} and XLM-RoBERTa \cite{9}. These methods learn statistical correspondences between character patterns and nationalities from training data, but struggle to generalize to patterns absent from the training data or to low-frequency nationalities. The second approach is direct reasoning using LLMs. Prior work \cite{10} demonstrated that having LLMs analyze the linguistic and cultural features of names substantially outperforms neural models. However, since LLMs attempt to apply abstract linguistic rules such as ``-ovich is a patronymic suffix typical of Slavic languages,'' they have limitations in fine-grained discrimination among multiple nationalities sharing the same pattern (e.g., Russian, Ukrainian, Serbian).

We hypothesize that nationality and region prediction performance can be further improved by more effectively eliciting LLM world knowledge. Our key observation is that LLMs exhibit higher reliability in concrete factual knowledge than in abstract linguistic rules. While an LLM may struggle to accurately explain ``the linguistic origins of -ovich,'' it can relatively easily recall famous individuals named Ivan Petrovich. This observation aligns with the cognitive processes humans employ when inferring the nationality of an unfamiliar name. For instance, upon encountering the name ``Tanaka,'' humans do not perform phonological analysis but rather recall famous individuals with the same name, such as Kakuei Tanaka or Shozo Tanaka, and infer the nationality from the fact that they are Japanese. In other words, by utilizing LLM world knowledge as associative memory for concrete examples rather than as a source for abstract reasoning, more effective knowledge elicitation becomes possible.

Based on this insight, we propose LLM Associative Memory Agents (LAMA). LAMA performs nationality prediction through indirect reasoning by recalling famous individuals with the same name as the input and aggregating their nationality information. Specifically, two agents specialized in different knowledge domains, a Person Agent that recalls general famous individuals and a Media Agent specialized in sports and entertainment, recall famous individuals in parallel, and Top-1 predictions are determined through voting based on the nationalities of recalled individuals. When recall fails, the system falls back to conventional direct prediction, and additional completion processing is applied for Top-K predictions. This approach resolves several challenges that were difficult for conventional methods. First, it becomes possible to distinguish nationalities within the same linguistic sphere that cannot be differentiated by abstract linguistic rules (e.g., Russian vs. Ukrainian, Argentine vs. Uruguayan) based on the existence of specific famous individuals of those nationalities. Second, even for low-frequency nationalities in the data, accurate prediction becomes possible if famous individuals of that nationality are contained in the LLM's world knowledge, mitigating the effects of data imbalance. Third, aggregation through voting from multiple famous individuals yields more stable predictions than conventional methods that depend on a single reasoning path, improving robustness against the stochastic output variation of LLMs. In essence, LAMA embodies a new multi-agent system that retrieves and aggregates LLM knowledge rather than prompting reasoning, enabling more direct and reliable elicitation of LLM world knowledge.

The main contributions of this study are as follows:

\begin{itemize}
    \item[i)] LAMA, a novel attribute prediction framework that leverages LLM world knowledge as associative memory, enables more reliable elicitation of concrete factual knowledge compared to abstract reasoning through indirect reasoning that recalls famous individuals with the same name and aggregates their nationalities rather than directly inferring nationality from names.
    
    \item[ii)] The dual-agent architecture comprising a Person Agent and a Media Agent specialized in different knowledge domains covers diverse famous individuals across fields that would be difficult to recall with a single agent, improving recall coverage and prediction robustness.
    
    \item[iii)] The combination of voting based on recall results with conditional completion processing achieves both Top-1 prediction accuracy and Top-K prediction diversity, allowing users to prioritize either precision or candidate coverage depending on their objectives.
\end{itemize}

The remainder of this paper is organized as follows. Section 2 reviews related tasks. Section 3 describes the details of our proposed method LAMA. Section 4 reports experiment settings and results, presenting multifaceted analyses. Section 5 discusses the key findings and limitations of this study. Section 6 concludes the paper.


\section{Related Work}

This section provides an overview of existing research on name-based attribute prediction and research on knowledge elicitation from LLMs.

\subsection{Name-based Attribute Prediction}

Research on predicting attributes such as nationality and ethnicity from personal names has been pursued for many years across the fields of social science, demography, and information science. Early research primarily employed statistical and rule-based methods. Mateos \cite{11} comprehensively reviewed 13 studies on name-based ethnic classification, reporting sensitivity ranges of 0.67--0.95 and specificity ranges of 0.8--1.0. Treeratpituk and Giles \cite{6} constructed an ethnic classification model using multinomial logistic regression from Wikipedia data, achieving 85\% accuracy. Voicu \cite{12} proposed the BIFSG model that utilizes first names in addition to surnames and geographic information, demonstrating superior accuracy compared to the conventional BISG method.

With the advancement of neural networks, deep learning methods were introduced to name-based attribute prediction. Ye et al. \cite{13} proposed a model using name embeddings learned from 57 million contact lists, achieving an F1 score of 0.795 for classification across 39 nationalities. Lee et al. \cite{14} proposed a method that directly predicts nationality from name strings using RNNs and evaluated it on Olympic athlete data. Hur \cite{15} classified Malaysian multi-ethnic names using LSTM and character embeddings, achieving an average accuracy improvement of 7\% over conventional methods. Xie \cite{7} developed the rethnicity package using Bi-LSTM models, improving prediction accuracy for minority ethnicities in U.S. voter data. These studies demonstrated that neural networks can automatically learn complex patterns in names, but achieving sufficient performance requires large amounts of labeled data, and generalization to patterns not present in the training data remained a challenge.

Recently, research on name-based attribute estimation using LLMs has become active. An et al. \cite{16} demonstrated that LLMs exhibit race, ethnicity, and gender bias in hiring decisions through name manipulation. Sakunkoo and Sakunkoo \cite{17} analyzed the tendency of LLMs to infer social status, race, and gender from names, visualizing implicit hierarchical biases. Phonchai et al. \cite{18} proposed Prompt-Engineered Fine-Tuning with cultural knowledge graph integration for LLMs, achieving 93.1\% accuracy in zero-shot multicultural name recognition. Their method improves the ability to infer cultural origins for unseen names by leveraging adversarial data augmentation and cultural knowledge graphs. Inoshita \cite{10} comprehensively compared neural models and LLM prompting methods, demonstrating that LLMs substantially outperform neural models and that this performance gap is attributable to world knowledge acquired during pre-training.

These studies indicate that LLMs possess rich knowledge about the relationships between names and attributes. However, the method of Phonchai et al. \cite{18} requires fine-tuning and external knowledge graphs, and conventional LLM prompting methods including Inoshita \cite{10} remain within the framework of directly inferring nationality from names. In this study, we explore methods for more effectively utilizing LLMs' internal knowledge without requiring additional training or external knowledge bases.

\subsection{Knowledge Elicitation from LLMs}

Methods for effectively eliciting knowledge from LLMs have been actively researched in recent years. CoT prompting \cite{2} is a technique that improves performance on complex reasoning tasks by having LLMs output their reasoning process step by step. Wei et al. demonstrated that CoT brings substantial performance improvements in arithmetic reasoning and commonsense reasoning. Self-Consistency \cite{19} is a method that generates multiple reasoning paths for the same problem and determines the final answer by majority voting, showing more stable results than depending on a single reasoning path. Self-Reflection \cite{3} is a technique that has LLMs evaluate and correct their own outputs, improving accuracy by self-correcting errors in initial answers.

In addition to these methods, approaches inspired by human cognitive processes have also been proposed. Chen et al. \cite{20} proposed Schema Activated ICL based on human schema theory, which explicitly activates abstract cognitive structures during LLM reasoning, improving zero-shot reasoning performance by up to 36\%. Budagam et al. \cite{21} proposed Hierarchical Prompting Taxonomy, which organizes prompts by cognitive load levels based on human cognitive hierarchy theory, reporting performance improvements of up to 63\%. Sumers et al. \cite{22} proposed Cognitive Architectures for Language Agents, a design framework for modular LLM agents that mimics human cognitive systems, achieving behavior closer to human mental processes by separating memory, action selection, and decision-making.

Multi-agent approaches that distribute and utilize LLM knowledge across multiple agents have also attracted attention. Kirk and Laird \cite{23} integrated LLMs into the Soar cognitive architecture, confirming that LLMs can function as an ``external cerebral cortex for knowledge bases.'' Shan et al. \cite{24} modeled LLM memory mechanisms by separating them into sensory, short-term, and long-term memory, proposing an external memory integration design that mimics human memory structures. Inoshita and Mizuno \cite{25} formulated sarcasm understanding as a human world model-like reasoning process (observation $\rightarrow$ latent state estimation $\rightarrow$ prediction $\rightarrow$ prediction error $\rightarrow$ intention judgment) and implemented each step by decomposing it into multiple LLM agents, achieving higher accuracy and interpretability than single models. Their research demonstrates the effectiveness of utilizing LLM world knowledge through multiple specialized agents.

LAMA in this study shares design philosophy with these cognitive science-inspired approaches. LAMA reproduces with LLMs one of the cognitive processes humans use when inferring the nationality of an unfamiliar name—recalling famous individuals with the same name and inferring from their attributes. However, LAMA is unique in that it directly utilizes LLM world knowledge as associative memory rather than mimicking abstract reasoning processes. That is, we propose a new knowledge utilization method that requests recall of concrete examples rather than having LLMs reason, and aggregates those results.


\section{Attribute Prediction Framework Using Associative Memory in LLMs}

\subsection{Framework Overview}

In this study, we propose LAMA, a novel framework that leverages the world knowledge acquired by LLMs through pre-training in a manner that mimics human associative memory processes. Conventional nationality prediction methods have adopted discriminative approaches that directly predict nationality labels from input names. In contrast, LAMA adopts an indirect reasoning approach inspired by the natural human cognitive process of recalling famous individuals with the same name and inferring from their attributes. The core insight of this framework is that LLMs, through training on large-scale text corpora, have internalized rich knowledge about numerous famous individuals' names and their attributes (nationality, occupation, field of activity, etc.). This knowledge can be elicited as associative memory through appropriate prompt design, providing a qualitatively different reasoning foundation compared to conventional methods that learn statistical patterns between names and attributes.

We begin by formally defining the input-output relationship of LAMA. Given an input personal name $n \in \mathcal{N}$, LAMA outputs an ordered sequence of attribute labels $\hat{\mathbf{y}} = (\hat{y}_1, \hat{y}_2, \ldots, \hat{y}_K)$, where $\hat{y}_k \in \mathcal{Y}$ represents the $k$-th ranked predicted label, and $\mathcal{Y}$ is the set of target attribute labels. In this study, we address three granularity levels for $\mathcal{Y}$: a set of 99 nationality labels, a set of 14 region labels, and a set of 6 continent labels. This capability to handle hierarchical attribute prediction tasks demonstrates the generality of the LAMA framework and ensures extensibility to broader applications of estimating personal attributes from names.

The LAMA processing pipeline consists of three main phases: the Associative Recall Phase, the Memory Aggregation Phase, and the Conditional Prediction Phase. In the Associative Recall Phase, two specialized agents operate in parallel to recall individuals associated with the input name. In the Memory Aggregation Phase, recall results from each agent are integrated, and occurrence frequencies are tallied for each attribute label. In the Conditional Prediction Phase, prediction strategies are adaptively switched based on the presence or absence of recall results, generating the final Top-K predictions. Through these three stages of processing, LAMA extracts LLM world knowledge in a structured manner and achieves reliable attribute prediction.

\begin{figure}[t]
\centering
\makebox[\textwidth][c]{%
  \includegraphics[width=1.30\textwidth]{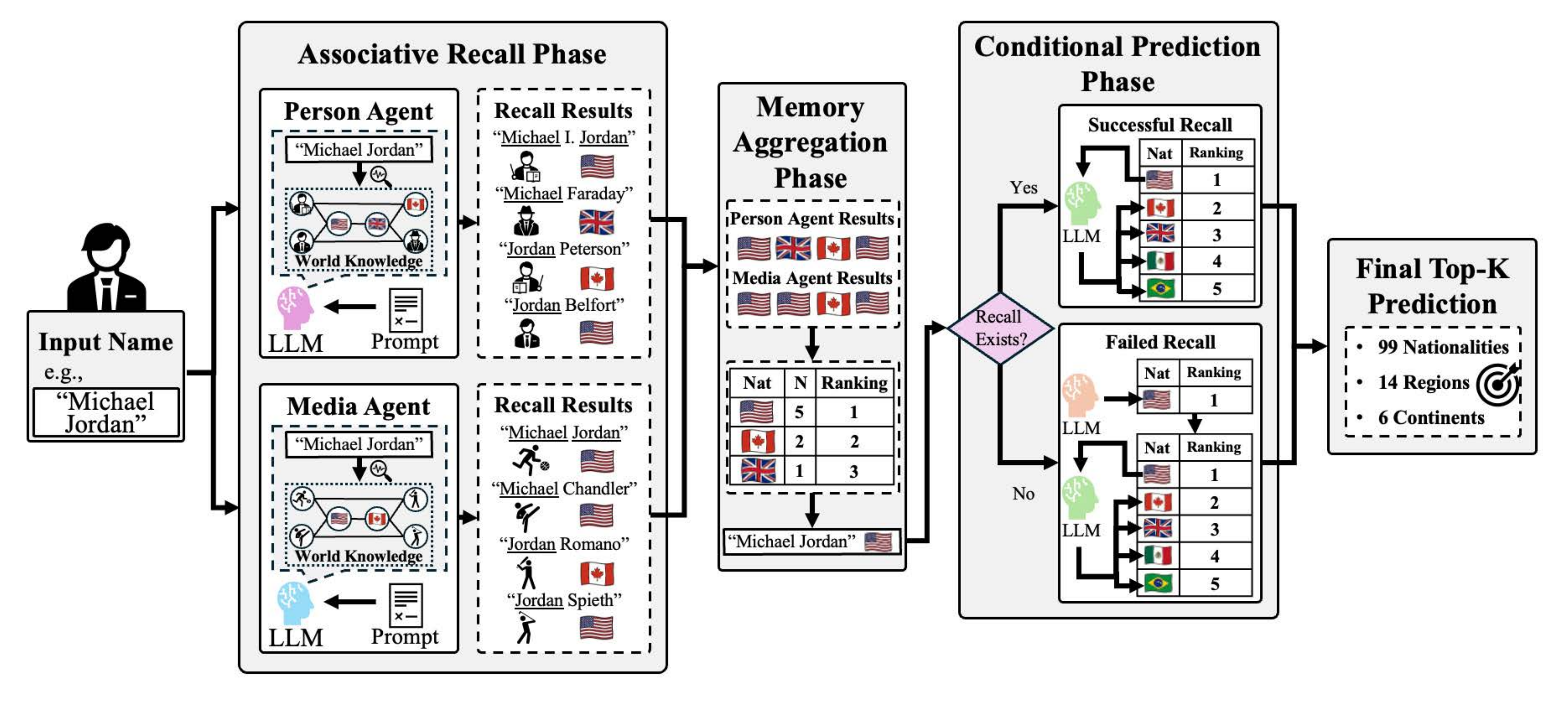}%
}
\caption{Overall architecture of the LAMA framework.}
\label{fig:framework}
\end{figure}

Fig. \ref{fig:framework} illustrates the overall architecture of LAMA. The input name is first sent to two parallel agents (Person Agent and Media Agent), and each agent independently recalls relevant individuals. The recalled individuals and their attributes are integrated in the aggregation module, generating a voting-based ranking. In the final stage, conditional branching is performed based on the existence of recall results, and Top-K predictions are output in combination with appropriate completion strategies.

\subsection{Dual Associative Memory Agents}

In the Associative Recall Phase of LAMA, we adopt a dual-agent architecture that operates two specialized agents (Person Agent and Media Agent) in parallel, rather than using a single agent. The rationale for this design is based on the observation that the recognizability and recallability of famous individuals varies significantly across domains. Individuals who are famous as politicians or scientists and those who are famous as athletes or entertainers may have different linguistic and cultural characteristics in their names, making it difficult for a single prompt to comprehensively cover both domains. The dual-agent configuration enables simultaneous achievement of specialization and comprehensiveness in recall for each domain.

Each agent is implemented as a single API call to an LLM, receiving an input name $n$ and recalling up to $M$ real famous individuals who have the same or very similar name, outputting pairs of each individual and their attributes. Here, similarity primarily refers to matching surnames or given names, judged based on the LLM's linguistic and cultural knowledge. This recall process enables indirect reasoning that estimates the attributes of the input name's owner from the attribute distribution of famous individuals with similar name patterns.

In this study, we set $M = 4$. This value was empirically selected considering the balance between the practical upper limit of individuals that LLMs can recall with high confidence and computational cost. Optimization of $M$ remains as future work, but experiment results demonstrate that the current setting achieves sufficient performance. Formally, the recall function $\mathcal{R}_a: \mathcal{N} \rightarrow 2^{\mathcal{P} \times \mathcal{Y}}$ of agent $a \in \{{\rm person}, {\rm media}\}$ returns a set of pairs of famous individuals' full names and attribute labels who have the same or similar name to input name $n$:
\begin{equation}
\mathcal{R}_a(n) = \{(p_1, y_1), (p_2, y_2), \ldots, (p_{m_a}, y_{m_a})\}
\end{equation}
where $m_a \leq M$ is the actual number of recalled individuals, $p_i \in \mathcal{P}$ is the full name of the recalled individual, and $y_i \in \mathcal{Y}$ is that individual's attribute label. Each agent is instructed to generate output in JSON array format, enabling structured information extraction.

\subsubsection{Person Agent}

The Person Agent is responsible for recalling individuals who have become famous primarily through social and intellectual activities, such as politicians, scientists, business leaders, and historical figures. The design philosophy of this agent is to elicit the ``general knowledge about famous individuals'' that LLMs have acquired through pre-training as associative memory.

The system prompt for the Person Agent is structured as follows:

\begin{quote}
\texttt{You are recalling real people based on a given name.}\\
\texttt{Think of REAL, ACTUAL famous people who have this exact name or a very similar name.}\\
\\
\texttt{For each person:}\\
\texttt{1. Full name}\\
\texttt{2. Nationality (from valid list only)}\\
\\
\texttt{Valid nationalities: [list of 99 nationality labels]}\\
\\
\texttt{Output JSON array of up to 4 people:}\\
\texttt{[\{"name": "Full Name", "nationality": "Nationality"\}]}\\
\\
\texttt{Be honest - only include people you are CONFIDENT actually exist.}
\end{quote}
This prompt design reflects several core design principles of the LAMA framework.

The first design principle is the explicit definition of the recall task. The opening instruction ``You are recalling real people'' clearly instructs the LLM to perform a recall task rather than a generation task. This causes the LLM to operate in a mode of searching and extracting relevant information from existing knowledge acquired during pre-training, rather than creating new information. This distinction is fundamentally important for ensuring output reliability.

The second design principle is appropriate scope setting for recall targets. The specification ``this exact name or a very similar name'' includes not only individuals with exact name matches but also those with matching surnames or given names, or those with culturally equivalent name variations (e.g., Michael and Michel). This flexibility enables richer recall results by leveraging name spelling variations and cross-cultural name correspondences.

The third design principle is output structuring and constraints. By constraining nationality labels to a predefined list of 99 countries, the LLM is prevented from outputting arbitrary strings, ensuring consistency with subsequent aggregation processing. The JSON array format specification enables structured information extraction and improves parsing reliability.

The fourth design principle is explicit constraints for hallucination suppression. The instruction ``Be honest - only include people you are CONFIDENT actually exist'' aims to suppress the LLM's tendency to generate non-existent individuals (hallucination). LLMs tend to output information even with low confidence, but this constraint guides them to output only high-confidence recall results.

The user prompt adopts a concise format that presents the input name $n$:

\begin{quote}
\texttt{Name: [input name]}\\
\\
\texttt{Recall real people with this name.}
\end{quote}
This concise user prompt design is intentional. Since detailed instructions for the recall task have already been given in the system prompt, the user prompt presents only the input information, avoiding duplication of task instructions. This is expected to focus the LLM's attention on the input name, yielding more relevant recall results.

Output from the Person Agent is returned as a JSON array. Each element contains the recalled individual's full name and nationality. The output JSON array is parsed, and nationality labels are normalized to the valid label set $\mathcal{Y}$ through case-insensitive matching. Entries with nationalities not in the valid label set or missing required fields are excluded, and up to $M=4$ valid recall results are retained.

\subsubsection{Media Agent}

The Media Agent is an agent specialized in recalling individuals who have become famous primarily through media exposure, such as athletes, actors, singers, and entertainers. The fundamental reason for designing this agent separately from the Person Agent is based on the observation that famous individuals in the media and entertainment field form a unique structure in LLMs' knowledge space.

The system prompt for the Media Agent is structured as follows:

\begin{quote}
\texttt{You are recalling athletes and entertainers based on a given name.}\\
\texttt{Think of REAL people from sports, movies, music, or TV who have this exact name or similar.}\\
\\
\texttt{For each person:}\\
\texttt{1. Full name}\\
\texttt{2. Nationality (from valid list only)}\\
\\
\texttt{Valid nationalities: [list of 99 nationality labels]}\\
\\
\texttt{Output JSON array of up to 4 people:}\\
\texttt{[\{"name": "Full Name", "nationality": "Nationality"\}]}\\
\\
\texttt{Be honest - only include people you are CONFIDENT actually exist.}
\end{quote}
The essential difference from the Person Agent lies in the task definition at the beginning of the prompt. The explicit limitation of targets to ``athletes and entertainers'' and the specific enumeration of fields as ``sports, movies, music, or TV'' focus the LLM's recall scope on the media and entertainment domain. This field limitation is not merely a scope constraint but a design that enables access to different regions of the LLM's knowledge space.

The reasons for designing the Media Agent as an independent agent are based on the following three observations. First, famous individuals in the media field tend to be difficult to recall with generic prompts. When using a generic prompt without field limitations, individuals such as politicians or historical figures tend to be preferentially recalled, and athletes or entertainers with equal or greater fame are often not recalled. This is thought to be due to exposure frequency in LLM pre-training data, inter-field biases, or response tendencies to prompts. By explicitly specifying the field, this bias can be corrected, ensuring that famous individuals in the media field are included in recall targets.

Second, nationality information of athletes and entertainers tends to be clearer and more reliable compared to other fields. Athletes have their nationality officially recorded through participation records in international competitions, and information about entertainers' birthplaces and activity bases is widely reported in the media. Due to this characteristic, recall results from the Media Agent are expected to contain high-accuracy nationality information.

Third, the globalization of the entertainment industry has formed structures where individuals of specific nationalities have international recognition. Examples include American actors in the Hollywood film industry, Korean artists in K-POP, and Brazilian or Argentine players in soccer. Such field-specific nationality distribution patterns differ from the politics and academic fields covered by the Person Agent and can only be effectively captured through the Media Agent.

The user prompt for the Media Agent adopts the same format as the Person Agent. This maintains consistency in input processing between the two agents and ensures comparability of recall results. The output format and parsing process are also identical to the Person Agent, and recall results from both agents are passed to the subsequent Memory Aggregation Phase in a unified format.

\subsubsection{Design Principles of Dual-Agent Architecture}

The two agents operate completely in parallel and are implemented as independent processes that do not reference each other's recall results. Specifically, API calls to both agents are issued asynchronously for input name $n$, and their responses are awaited in parallel. Due to this parallel execution, the two agents are processed simultaneously, so the latency of the recall phase is effectively equivalent to that of a single agent. This achieves increased recall capacity from additional agents without increased execution time.

The reasons for adopting the dual-agent configuration are the following three points. First, the recognizability and recallability of famous individuals vary significantly across domains. Individuals who are famous as politicians or scientists and those who are famous as athletes or entertainers may have different representations in the LLM's knowledge space. Therefore, the generic Person Agent alone may not sufficiently recall famous individuals in the media field, and adding the field-specialized Media Agent enables recall from a wide range of domains.

Second, when the same individual is redundantly recalled by both agents, it serves as a strong signal that the individual is famous in multiple fields. For example, individuals who transitioned from actor to politician or individuals known both as athletes and business leaders may be recalled by both agents. This redundancy is reflected as increased vote counts for the corresponding attribute labels in the subsequent Memory Aggregation Phase, contributing to improved prediction reliability.

Third, the complementarity of the two agents makes it possible to cover name patterns that cannot be captured by a single agent. If a specific name is associated with famous individuals only in one field, recall may fail without an agent specialized in that field. The dual-agent configuration has the effect of improving such recall comprehensiveness.

Agent independence is also an important design principle. Since each agent operates independently without referencing the other's recall results, one agent's output does not influence the other. This independence allows each agent to focus on recall within its specialized domain, resulting in diverse recall results. Additionally, independent processing enables parallel execution, contributing to improved computational efficiency.

Note that extension to three or more agents is possible in principle, but in this study, we adopt a two-agent configuration considering the trade-off between the number of agents and API call costs. Furthermore, regarding such complexity in prediction processes, Inoshita \cite{10} argued that increasing prompt complexity does not necessarily improve accuracy, and this study follows that consideration by limiting the configuration to two agents.

Recall results from each agent are integrated in the subsequent Memory Aggregation Phase and utilized as a comprehensive evidence set. The effectiveness of the dual-agent configuration, namely that both agents contribute complementarily to prediction accuracy, is quantitatively verified through comparison with conditions where each agent is individually removed in the ablation analysis in the experiments section.

\subsection{Memory Aggregation and Voting Mechanism}

Recall results from each agent obtained in the Associative Recall Phase are integrated in the Memory Aggregation Phase and utilized as evidence for attribute prediction. The purpose of this phase is to convert heterogeneous recall results from multiple agents into a single consistent ranking.

\subsubsection{Integration of Recall Results}

Recall results from the two agents are integrated as a simple union. Letting the Person Agent's recall results be $\mathcal{R}_{\rm person}(n)$ and the Media Agent's recall results be $\mathcal{R}_{\rm media}(n)$, the integrated recall set $\mathcal{R}(n)$ is defined by the following equation:
\begin{equation}
\mathcal{R}(n) = \mathcal{R}_{\rm person}(n) \cup \mathcal{R}_{\rm media}(n)
\end{equation}
In this integration, the same individual may be redundantly recalled by both agents, but this study intentionally allows this. This is because redundant recall of the same individual is a strong signal that the individual is famous in multiple fields, functioning as evidence that increases the reliability of the corresponding attribute label. The size of the integrated recall set is $|\mathcal{R}(n)| \leq 2M$, and in this study's setting, up to 8 individuals may be recalled.

From the integrated recall set, occurrence frequencies for each attribute label are counted. The frequency count $C(y)$ for attribute label $y \in \mathcal{Y}$ is calculated by the following equation:
\begin{equation}
C(y) = \sum_{(p, y') \in \mathcal{R}(n)} \mathbbm{1}[y' = y]
\end{equation}
where $\mathbbm{1}[\cdot]$ is the indicator function that returns 1 when the condition is true and 0 when false. This frequency count represents the empirical distribution of attribute labels among recalled individuals and forms the basis for prediction based on the majority voting principle.

\subsubsection{Voting-based Ranking Generation}

Based on frequency counts, an ordered list of attribute labels is generated. The first-ranked prediction label $\hat{y}_1$ is determined as the most frequently occurring attribute label:
\begin{equation}
\hat{y}_1 = \arg\max_{y \in \mathcal{Y}} C(y)
\end{equation}
In case of ties, the decision is based on the recall order. This majority voting-based prediction has an ensemble-like effect of integrating information from multiple independent pieces of evidence (recalled individuals), enabling robust prediction against noise potentially contained in individual recall results.

Defining the set of attribute labels with positive frequency counts as $\mathcal{Y}^+ = \{y \in \mathcal{Y} : C(y) > 0\}$, these labels are preferentially considered as candidates with direct evidence from recall in subsequent prediction generation. On the other hand, labels with $C(y) = 0$ have no evidence from recall but can be considered in the completion phase.

\subsection{Conditional Prediction Strategy}

In the final phase of LAMA, we adopt a conditional prediction mechanism that adaptively switches prediction strategies based on the presence or absence of recall results. This design is based on the observation that optimal prediction strategies differ between successful and failed recall cases.

\subsubsection{Prediction Strategy on Successful Recall}

When the recall set is not empty ($|\mathcal{R}(n)| \geq 1$), LAMA adopts a strategy that maximally utilizes evidence from recall. Specifically, the first-ranked prediction $\hat{y}_1$ determined by voting is fixed, and the remaining ranks (2nd through $K$th) are generated by LLM completion.

In completion processing, the LLM is presented with the input name $n$, the list of recalled individuals $\mathcal{R}(n)$, and the confirmed first-ranked prediction $\hat{y}_1$, and is instructed to generate candidates from the second rank onwards. The system prompt used for completion processing is structured as follows:

\begin{quote}
\texttt{Given a name and the most likely nationality (rank 1), suggest 4 more nationalities that could also be possible.}\\
\\
\texttt{Consider:}\\
\texttt{1. Culturally/geographically similar countries}\\
\texttt{2. Countries where this name pattern might also appear}\\
\texttt{3. Historical migration patterns}\\
\\
\texttt{The rank 1 nationality is already determined. Suggest ranks 2-5.}\\
\\
\texttt{Valid nationalities: [list of 99 nationality labels]}\\
\texttt{Output a JSON array of exactly 4 nationalities for ranks 2-5.}
\end{quote}
This prompt design explicitly instructs reasoning from three perspectives—cultural and geographical relevance, name pattern occurrence distribution, and historical migration patterns—rather than mere enumeration of similar nationalities. This enables meaningful ordering even for candidates other than Top-1.

The user prompt for completion processing contains different information depending on the presence or absence of recall results. On successful recall, the following format is adopted:

\begin{quote}
\texttt{Name: [input name]}\\
\\
\texttt{Recalled people: [list of recalled individuals (JSON format)]}\\
\texttt{Rank 1 (from recall): [first-ranked prediction]}\\
\\
\texttt{Suggest 4 more nationalities for ranks 2-5.}
\end{quote}
By including the list of recalled individuals in the completion prompt, the LLM can perform completion considering the context based on recall results. For example, if recalled individuals have multiple different nationalities, that diversity is expected to be reflected in the completion results.

Let the completion output from the LLM be $\mathcal{C}(n, \hat{y}_1) = (c_1, c_2, \ldots, c_{K-1})$. The final Top-K prediction is generated by integrating the first-ranked prediction, completion results, and other attribute labels obtained from recall. Attribute labels that appeared in recall but are not the first rank, $\mathcal{Y}^+ \setminus \{\hat{y}_1\}$, are preferentially placed in higher ranks than completion results. This is based on the assumption that evidence from actual recall is more reliable than LLM's general reasoning. The final prediction is constructed by the following procedure:
\begin{equation}
\hat{\mathbf{y}} = (\hat{y}_1) \oplus \text{Unique}((\mathcal{Y}^+ \setminus \{\hat{y}_1\}) \oplus \mathcal{C}(n, \hat{y}_1))
\end{equation}
where $\oplus$ represents sequence concatenation and $\text{Unique}(\cdot)$ is an operation that removes duplicates while preserving order. With this strategy, prediction is completed with an average of 3 API calls (2 parallel recalls + 1 completion) on successful recall.

\subsubsection{Prediction Strategy on Failed Recall}

When the recall set is empty ($|\mathcal{R}(n)| = 0$), a different prediction strategy is required since no evidence from recall is available. This situation occurs when the input name is very rare, or conversely, too common to be associated with specific famous individuals. In the experiment data, approximately 17--18\% of samples fall into this case.

On failed recall, LAMA adopts a two-stage direct prediction strategy. In the first stage, the LLM is instructed to directly predict Top-K attributes from the input name, obtaining initial predictions $\hat{\mathbf{y}}^{\rm init} = (\hat{y}_1^{\rm init}, \ldots, \hat{y}_K^{\rm init})$. The system prompt used for direct prediction is as follows:

\begin{quote}
\texttt{You are an expert in identifying the nationality of people based on their names.}\\
\texttt{Predict the TOP 5 most likely nationalities for the given name.}\\
\\
\texttt{Valid nationalities: [list of 99 nationality labels]}\\
\texttt{Output a JSON array of 5 nationalities.}
\end{quote}
This prompt is essentially equivalent to conventional LLM prompting methods (zero-shot prediction). The accuracy of zero-shot prediction in this case is guaranteed by Inoshita's investigation \cite{10}. Following this method, the LLM reasons nationality directly from the linguistic and cultural features of the input name. The user prompt adopts a concise format:

\begin{quote}
\texttt{Name: [input name]}
\end{quote}

In the second stage, the first-ranked prediction $\hat{y}_1 = \hat{y}_1^{\rm init}$ is fixed, and completion processing similar to successful recall is applied to regenerate the second rank onwards. In the completion prompt for failed recall, since no recall results exist, the user prompt takes the following simplified form:

\begin{quote}
\texttt{Name: [input name]}\\
\texttt{Rank 1 (confirmed): [first-ranked prediction]}\\
\\
\texttt{Suggest 4 more nationalities for ranks 2-5.}
\end{quote}
The purpose of this two-stage processing is to improve the reliability of direct prediction. In preliminary experiments, a tendency for prediction quality to decline for lower ranks was observed when generating the entire Top-K in a single API call. This is thought to be due to the LLM's tendency to focus attention on higher ranks and not perform sufficient reasoning for lower ranks. By fixing the first rank and separately executing completion processing, it becomes possible to maintain consistency of the first-ranked prediction while improving diversity and relevance of lower ranks.

On failed recall, a total of 4 API calls (2 parallel recalls (empty results) + 1 direct prediction + 1 completion) are required. One more API call is required compared to successful recall, but this additional cost is tolerated to maintain prediction quality for the edge case of failed recall. In general, prediction accuracy on failed recall is expected to decline compared to successful recall, but this fallback mechanism enables LAMA to output predictions for all inputs.

\subsection{Overall Algorithm}

This section presents the complete algorithm of LAMA as pseudocode and analyzes computational complexity and API call counts. Algorithm \ref{alg:lama} shows the overall processing flow of LAMA. The algorithm takes a name $n$ and output rank count $K$ as input and returns Top-K attribute predictions $\hat{\mathbf{y}}$.

\begin{algorithm}[t]
\caption{LAMA: LLM Associative Memory Agents}
\label{alg:lama}
\begin{algorithmic}[1]
\Require Input name $n$, output rank count $K$, maximum recall count $M$
\Ensure Top-K attribute predictions $\hat{\mathbf{y}} = (\hat{y}_1, \ldots, \hat{y}_K)$

\Statex \textbf{// Phase 1: Associative Recall}
\State $\mathcal{R}_{\rm person} \gets \text{PersonAgent}(n, M)$ \Comment{Parallel execution}
\State $\mathcal{R}_{\rm media} \gets \text{MediaAgent}(n, M)$ \Comment{Parallel execution}
\State $\mathcal{R} \gets \mathcal{R}_{\rm person} \cup \mathcal{R}_{\rm media}$

\Statex

\Statex \textbf{// Phase 2: Memory Aggregation}
\For{each $y \in \mathcal{Y}$}
    \State $C(y) \gets \sum_{(p, y') \in \mathcal{R}} \mathbbm{1}[y' = y]$
\EndFor

\Statex

\Statex \textbf{// Phase 3: Conditional Prediction}
\If{$|\mathcal{R}| = 0$}
    \State $\hat{\mathbf{y}}^{\rm init} \gets \text{DirectPredict}(n, K)$
    \State $\hat{y}_1 \gets \hat{y}_1^{\rm init}$
\Else
    \State $\hat{y}_1 \gets \arg\max_{y} C(y)$
\EndIf

\Statex

\Statex \textbf{// Phase 4: Top-K Completion}
\State $\mathcal{C} \gets \text{Complete}(n, \mathcal{R}, \hat{y}_1, K-1)$
\State $\mathcal{Y}^+ \gets \{y : C(y) > 0\} \setminus \{\hat{y}_1\}$
\State $\hat{\mathbf{y}} \gets (\hat{y}_1) \oplus \text{Unique}(\mathcal{Y}^+ \oplus \mathcal{C})$

\Statex

\State \Return $\hat{\mathbf{y}}[1:K]$
\end{algorithmic}
\end{algorithm}

The computational complexity of LAMA is primarily determined by the number of LLM API calls. Since the Person Agent and Media Agent are executed in parallel, the latency of the recall phase is effectively equivalent to a single agent call, resulting in a scalable design with respect to the increase in the number of agents. On successful recall ($|\mathcal{R}| \geq 1$), a total of 3 calls are required, and on failed recall ($|\mathcal{R}| = 0$), a total of 4 calls are required. Considering that the recall success rate in the experiment data is approximately 82\%, the expected number of API calls per sample is $0.82 \times 3 + 0.18 \times 4 = 3.18$.

The computational complexity of the Memory Aggregation Phase is $O(|\mathcal{R}| \cdot |\mathcal{Y}|)$, but since $|\mathcal{R}| \leq 2M$ and $|\mathcal{Y}|$ is constant (99 nationalities or 14 regions), it can be effectively considered as $O(1)$. Therefore, the overall computational complexity of LAMA is dominated by API call waiting time, and efficient inference is possible through parallel execution.

The LAMA framework described in this chapter extracts LLM world knowledge in a structured manner as associative memory and performs attribute prediction through indirect reasoning that mimics human cognitive processes. Through the combination of diverse recall via dual-agent configuration, reliable aggregation based on voting, and adaptive processing via conditional prediction strategies, it achieves performance surpassing conventional direct prediction methods.


\section{Experiment and Analysis}

\subsection{Experiment Design}

This section describes the dataset used to evaluate the proposed method LAMA, the baseline methods for comparison, and the evaluation metrics.

\subsubsection{Dataset}

In this study, we adopt the same dataset and preprocessing procedures as prior work \cite{10}, using a nationality prediction dataset constructed based on the name2nat dataset \cite{26}. The name2nat dataset is a large-scale corpus containing personal names and corresponding nationality labels, covering 890,248 romanized personal names and nationality information from 173 countries.

The following preprocessing steps were applied to the original name2nat dataset. First, to ensure sufficient training samples for each nationality, only nationalities with 500 or more samples were selected. This threshold was set as the minimum number of samples necessary for models to acquire nationality-specific features. This processing reduced the number of target nationalities from 173 to 99. Second, to mitigate severe class imbalance, the maximum number of samples per nationality was set to 800, with excess samples excluded through random sampling. This upper limit was determined from the perspective of suppressing excessive influence from majority classes on learning while maintaining sufficient data scale. Third, data was split into training, validation, and test sets at an 8:1:1 ratio based on stratified sampling. The application of stratified sampling ensured that all nationalities were included in appropriate proportions in each split, and random shuffling was performed after splitting. After these preprocessing steps, the final dataset consists of 75,345 samples and 99 nationality classes. Table \ref{tab:dataset} shows the breakdown of the dataset. Training and validation data were used for training the neural models used as baselines. On the other hand, since LLM prompting methods and the proposed method LAMA do not require additional training, evaluation was performed using only the test data. This evaluation design ensures that all methods are compared under identical conditions on the test data.

\begin{table}[t]
    \centering
    \caption{Dataset overview}
    \label{tab:dataset}
    \begin{tabular}{lcccc}
        \hline
        & Train & Validation & Test & Total \\
        \hline
        Samples & 60,277 & 7,534 & 7,534 & 75,345 \\
        \hline
        Nationality classes & \multicolumn{4}{c}{99} \\
        Region classes & \multicolumn{4}{c}{14} \\
        Continent classes & \multicolumn{4}{c}{6} \\
        \hline
    \end{tabular}
\end{table}

To enable evaluation at different granularity levels, the 99 nationalities were mapped to 14 regional categories and 6 continental categories. Table \ref{tab:region} shows the mapping definition from 14 regions to 6 continents and the number of nationalities included in each region. This mapping was defined considering geographical proximity and cultural and linguistic commonalities. As specific examples, nationalities from East Asian countries share romanization patterns derived from Chinese characters, and nationalities from Western European countries have Latin or Germanic naming conventions. The mapping to 6 continents was defined by further aggregating the 14 regions. Asia includes four regions: East Asia, Southeast Asia, South Asia, and Caucasus \& Central Asia. Europe includes four regions: Western Europe, Northern Europe, Southern Europe, and Eastern Europe. Americas includes three regions: North America, Central America \& Caribbean, and South America. Middle East, Africa, and Oceania were each treated as single continental categories. This hierarchical region definition enables evaluation of model performance at three granularity levels: nationality (99 classes), region (14 classes), and continent (6 classes).

\begin{table}[t]
    \centering
    \caption{Region definitions and hierarchical structure}
    \label{tab:region}
    \begin{tabular}{llr}
        \hline
        6 Continents & 14 Regions & Nationalities \\
        \hline
        \multirow{4}{*}{Asia} 
            & East Asia & 5 \\
            & Southeast Asia & 7 \\
            & South Asia & 6 \\
            & Caucasus \& Central Asia & 2 \\
        \hline
        \multirow{4}{*}{Europe} 
            & Western Europe & 11 \\
            & Northern Europe & 1 \\
            & Southern Europe & 5 \\
            & Eastern Europe & 15 \\
        \hline
        \multirow{3}{*}{Americas} 
            & North America & 3 \\
            & Central America \& Caribbean & 7 \\
            & South America & 10 \\
        \hline
        Middle East & Middle East & 10 \\
        \hline
        Africa & Africa & 15 \\
        \hline
        Oceania & Oceania & 2 \\
        \hline
        \multicolumn{2}{l}{Total} & 99 \\
        \hline
    \end{tabular}
\end{table}

\subsubsection{Baseline}

To evaluate the performance of the proposed method LAMA, we adopt both neural models and LLM prompting methods as baselines. These baselines were comprehensively compared and evaluated in prior work \cite{10}, and in this study, we perform comparisons using the same dataset and evaluation protocol.

As neural baselines, we adopt the following three models:

\begin{itemize}
    \item[i)] Support Vector Machine (SVM): A linear classifier using TF-IDF features of character n-grams. Despite being a simple method, it is known as a strong baseline for text classification. In this study, we use character 1-gram to 4-gram features.
    \item[ii)] CANINE \cite{8}: A Transformer model pre-trained at the character level. Since it accepts Unicode characters directly as input without using a tokenizer, it enables robust processing for multilingual text.
    \item[iii)] XLM-RoBERTa \cite{9}: A multilingual Transformer model pre-trained on over 100 languages. Although it uses subword tokenization, training on multilingual corpora enables processing of names from various linguistic backgrounds. It achieves the highest performance among neural baselines.
\end{itemize}

As LLM baselines, we adopt the following three prompting methods. GPT-4.1-mini is used for all methods:

\begin{itemize}
    \item[iv)] CoT \cite{2}: A method that has the model output the reasoning process leading to the prediction step by step. It analyzes features of names (phonological patterns, suffixes, etymology, character combinations, etc.) and outputs predictions after reasoning about which nationalities/regions those features are related to.
    \item[v)] Self-Consistency \cite{19}: A method that performs multiple inferences for the same input and determines the final prediction by majority voting. In this study, 5 independent inferences are executed for the same name, and the most frequently predicted nationality is adopted as the final result.
    \item[vi)] Self-Reflection \cite{3}: A method that has the model self-evaluate its initial prediction and make corrections as needed. After making an initial prediction, the model is made to consider the confidence of that prediction and potential errors, and if confidence is low or alternative candidates exist, it outputs the final prediction after reconsideration.
\end{itemize}

These baselines cover diverse approaches including classical machine learning (SVM), pre-trained models (CANINE, XLM-RoBERTa), and LLM prompting (CoT, Self-Consistency, Self-Reflection), enabling comparisons that clarify the positioning of the proposed method. Additionally, the proposed method LAMA also uses GPT-4.1-mini, the same as the prompting methods, and the LLM temperature parameter was set to 1.0 for LAMA and all prompting methods.

\subsubsection{Evaluation Metrics}

Nationality and region prediction tasks have inherent characteristics that require consideration of class imbalance, task-specific ambiguity, and the closeness of predictions. Based on these challenges, this study uses the following evaluation metrics.

\begin{itemize}
    \item[i)] Accuracy: Represents the correct rate of Top-1 predictions, i.e., the proportion of samples where the most confident prediction matches the correct label. As the most fundamental evaluation metric, it is widely used for comparison with prior work.
    
    \item[ii)] Macro-F1: A metric that calculates F1 scores for each class and takes their arithmetic mean. Since it weights all classes equally, performance on classes with few samples is also evaluated fairly. For tasks with large imbalances in sample sizes between classes, such as nationality and region prediction, it enables more balanced evaluation compared to Accuracy. Macro-F1 is calculated by the following equation:
    \begin{equation}
        \text{Macro-F1} = \frac{1}{C} \sum_{i=1}^{C} \text{F1}_i = \frac{1}{C} \sum_{i=1}^{C} \frac{2 \cdot P_i \cdot R_i}{P_i + R_i}
    \end{equation}
    where $C$ is the total number of classes, and $P_i$ and $R_i$ are the precision and recall for class $i$, respectively.
    
    \item[iii)] Precision@K: A metric indicating the proportion of samples where the correct label is included in the Top-K predictions. In this study, we adopt $K = 3, 5$. Since there are cases where uniquely identifying nationality/region from names alone is inherently difficult, conducting Top-K evaluation in addition to Top-1 evaluation allows verification of whether models include the correct answer in their candidate sets. Precision@K is defined as follows:
    \begin{equation}
        \text{Precision@}K = \frac{1}{N} \sum_{j=1}^{N} \mathbbm{1}(y_j \in \text{Top}_K(\hat{y}_j))
    \end{equation}
    where $N$ is the total number of samples, $y_j$ is the correct label for sample $j$, $\text{Top}_K(\hat{y}_j)$ is the set of Top-K predictions for sample $j$, and $\mathbbm{1}(\cdot)$ is the indicator function.
\end{itemize}

In addition to the above metrics, we perform stratified analysis based on label frequency to verify model robustness. Specifically, nationalities are classified into three groups, Head (high frequency), Mid (medium frequency), and Tail (low frequency), according to their occurrence frequency in the training data, and performance is measured separately for each group. Through this analysis, we clarify how robust models are to low-frequency nationalities and whether frequency-dependent biases exist.

\subsection{Evaluation on Nationality Prediction}

Table \ref{tab:nationality_results} shows the performance of each method on the 99-country nationality classification task. Accuracy, Macro-F1 score, and Precision@K (K=3, 5) were used as evaluation metrics. For each metric, we report the mean and standard deviation of results from three runs with different random seeds.

\begin{table}[t]
\centering
\caption{Performance comparison of methods on the 99-country nationality classification task}
\label{tab:nationality_results}
\begin{tabular}{lcccc}
\toprule
Method & Accuracy & Macro-F1 & Precision@3 & Precision@5 \\
\midrule
\multicolumn{5}{l}{\textit{Neural Models}} \\
SVM & 0.481 {\scriptsize $\pm$ 0.004} & 0.466 {\scriptsize $\pm$ 0.005} & 0.644 {\scriptsize $\pm$ 0.006} & 0.710 {\scriptsize $\pm$ 0.005} \\
CANINE & 0.450 {\scriptsize $\pm$ 0.013} & 0.435 {\scriptsize $\pm$ 0.013} & 0.648 {\scriptsize $\pm$ 0.012} & 0.733 {\scriptsize $\pm$ 0.010} \\
XLM-RoBERTa & 0.446 {\scriptsize $\pm$ 0.010} & 0.426 {\scriptsize $\pm$ 0.009} & 0.647 {\scriptsize $\pm$ 0.007} & 0.732 {\scriptsize $\pm$ 0.002} \\
\midrule
\multicolumn{5}{l}{\textit{LLM Prompting Methods}} \\
CoT & 0.709 {\scriptsize $\pm$ 0.009} & 0.715 {\scriptsize $\pm$ 0.009} & 0.821 {\scriptsize $\pm$ 0.007} & 0.853 {\scriptsize $\pm$ 0.006} \\
Self-Consistency & 0.763 {\scriptsize $\pm$ 0.006} & 0.769 {\scriptsize $\pm$ 0.005} & 0.855 {\scriptsize $\pm$ 0.005} & 0.880 {\scriptsize $\pm$ 0.005} \\
Self-Reflection & 0.776 {\scriptsize $\pm$ 0.005} & 0.782 {\scriptsize $\pm$ 0.004} & 0.870 {\scriptsize $\pm$ 0.006} & 0.893 {\scriptsize $\pm$ 0.005} \\
\midrule
LAMA & 0.817 {\scriptsize $\pm$ 0.006} & 0.824 {\scriptsize $\pm$ 0.005} & 0.885 {\scriptsize $\pm$ 0.004} & 0.902 {\scriptsize $\pm$ 0.004} \\
\bottomrule
\end{tabular}
\end{table}

LAMA achieved the highest performance on all evaluation metrics. In terms of accuracy, LAMA achieved 0.817, surpassing Self-Reflection (0.776), the highest-performing LLM baseline method, by 0.041 points. Compared to SVM (0.481), the highest-performing neural model, this represents a substantial improvement of 0.336 points. A similar trend was observed for Macro-F1 score, with LAMA achieving 0.824, surpassing Self-Reflection (0.782) by 0.042 points.

The Precision@K results indicate that LAMA's predictions are high quality not only for Top-1 but also for other candidates. In Precision@3, LAMA achieved 0.885, surpassing Self-Reflection (0.870) by 0.015 points. In Precision@5, LAMA reached 0.902, with a difference of 0.009 points from Self-Reflection (0.893). The fact that the improvement margin for Precision@K is smaller compared to the accuracy improvement (0.041 points) suggests that LAMA's main contribution lies in improving Top-1 prediction accuracy. This is considered to be due to the ability of the recall-based voting mechanism to accurately identify the most confident prediction.

A notable performance gap exists between neural models and LLM-based methods. Even CoT (0.709), the simplest LLM method, surpasses SVM (0.481), the highest-performing neural model, by 0.228 points. This result indicates that for the task of predicting nationality from names, LLM world knowledge is more effective than statistical patterns learned from training data. Interestingly, the pre-trained models CANINE and XLM-RoBERTa showed lower performance than SVM, which is based on simple features. This suggests that general-purpose multilingual pre-training is not necessarily effective for specialized tasks like the correspondence between names and nationalities.

In comparison among LLM methods, advanced prompting methods such as Self-Consistency (0.763) and Self-Reflection (0.776) show performance exceeding CoT (0.709). LAMA further surpasses these methods, achieving an improvement of 0.041 points over Self-Reflection. Importantly, LAMA's improvement is not dependent on complex reasoning chains or multiple self-corrections, but is achieved through an intuitive approach based on human cognitive processes—recalling famous individuals. This result demonstrates that indirect reasoning that explicitly leverages LLM world knowledge is more effective than conventional direct nationality prediction.

From the perspective of standard deviation, all methods show relatively stable performance, and LAMA's standard deviation (Accuracy: 0.006) is comparable to other LLM methods. This indicates that LAMA's recall-based approach has stability equivalent to Self-Consistency, which depends on stochastic sampling.

\subsection{Evaluation on Region Prediction}

In addition to nationality prediction, we evaluate LAMA's performance on region prediction tasks at coarser granularities. Region prediction is important for practical applications because it can provide useful information based on cultural and geographical proximity even when accurate identification of nationality is difficult. In this study, we evaluated at two granularity levels: 14-region classification and 6-continent classification.

Table \ref{tab:region14_results} shows the results of the 14-region classification task. In 14-region classification, LAMA achieved the highest performance on all metrics. In terms of accuracy, LAMA achieved 0.857, surpassing Self-Reflection (0.751) by 0.106 points. This improvement margin is larger than the improvement margin in 99-country classification (0.041 points). In Macro-F1 as well, LAMA achieved 0.829, with a difference of 0.125 points from Self-Reflection (0.704). In Precision@2 and Precision@3, LAMA achieved 0.920 and 0.948, respectively, surpassing all other methods.

\begin{table}[t]
\centering
\caption{Performance comparison of methods on the 14-region classification task}
\label{tab:region14_results}
\begin{tabular}{lcccc}
\toprule
Method & Accuracy & Macro-F1 & Precision@2 & Precision@3 \\
\midrule
\multicolumn{5}{l}{\textit{Neural Models}} \\
SVM & 0.682 {\scriptsize $\pm$ 0.005} & 0.619 {\scriptsize $\pm$ 0.003} & 0.812 {\scriptsize $\pm$ 0.004} & 0.872 {\scriptsize $\pm$ 0.002} \\
CANINE & 0.683 {\scriptsize $\pm$ 0.006} & 0.621 {\scriptsize $\pm$ 0.001} & 0.824 {\scriptsize $\pm$ 0.003} & 0.888 {\scriptsize $\pm$ 0.003} \\
XLM-RoBERTa & 0.690 {\scriptsize $\pm$ 0.008} & 0.631 {\scriptsize $\pm$ 0.010} & 0.830 {\scriptsize $\pm$ 0.002} & 0.894 {\scriptsize $\pm$ 0.001} \\
\midrule
\multicolumn{5}{l}{\textit{LLM Prompting Methods}} \\
CoT & 0.697 {\scriptsize $\pm$ 0.001} & 0.650 {\scriptsize $\pm$ 0.003} & 0.859 {\scriptsize $\pm$ 0.001} & 0.905 {\scriptsize $\pm$ 0.001} \\
Self-Consistency & 0.710 {\scriptsize $\pm$ 0.001} & 0.659 {\scriptsize $\pm$ 0.005} & 0.876 {\scriptsize $\pm$ 0.001} & 0.914 {\scriptsize $\pm$ 0.002} \\
Self-Reflection & 0.751 {\scriptsize $\pm$ 0.004} & 0.704 {\scriptsize $\pm$ 0.005} & 0.891 {\scriptsize $\pm$ 0.001} & 0.919 {\scriptsize $\pm$ 0.002} \\
\midrule
LAMA & 0.857 {\scriptsize $\pm$ 0.005} & 0.829 {\scriptsize $\pm$ 0.004} & 0.920 {\scriptsize $\pm$ 0.002} & 0.948 {\scriptsize $\pm$ 0.001} \\
\bottomrule
\end{tabular}
\end{table}

A notable point is that in 14-region classification, the performance gap between neural models and LLM methods is reduced compared to nationality classification. The difference between XLM-RoBERTa (0.690) and CoT (0.697) is only 0.007 points. This suggests that superficial features of names (character patterns) function more effectively for region-level classification. However, LAMA still substantially surpasses all methods, and the superiority of the recall-based approach is maintained even as granularity becomes coarser.

\begin{table}[t]
\centering
\caption{Performance comparison of methods on the 6-continent classification task}
\label{tab:region6_results}
\begin{tabular}{lcccc}
\toprule
Method & Accuracy & Macro-F1 & Precision@2 & Precision@3 \\
\midrule
\multicolumn{5}{l}{\textit{Neural Models}} \\
SVM & 0.767 {\scriptsize $\pm$ 0.001} & 0.679 {\scriptsize $\pm$ 0.007} & 0.899 {\scriptsize $\pm$ 0.001} & 0.952 {\scriptsize $\pm$ 0.003} \\
CANINE & 0.776 {\scriptsize $\pm$ 0.003} & 0.696 {\scriptsize $\pm$ 0.009} & 0.906 {\scriptsize $\pm$ 0.002} & 0.959 {\scriptsize $\pm$ 0.002} \\
XLM-RoBERTa & 0.780 {\scriptsize $\pm$ 0.005} & 0.692 {\scriptsize $\pm$ 0.012} & 0.911 {\scriptsize $\pm$ 0.002} & 0.960 {\scriptsize $\pm$ 0.000} \\
\midrule
\multicolumn{5}{l}{\textit{LLM Prompting Methods}} \\
CoT & 0.827 {\scriptsize $\pm$ 0.005} & 0.753 {\scriptsize $\pm$ 0.003} & 0.940 {\scriptsize $\pm$ 0.002} & 0.978 {\scriptsize $\pm$ 0.001} \\
Self-Consistency & 0.824 {\scriptsize $\pm$ 0.002} & 0.760 {\scriptsize $\pm$ 0.008} & 0.952 {\scriptsize $\pm$ 0.001} & 0.987 {\scriptsize $\pm$ 0.001} \\
Self-Reflection & 0.847 {\scriptsize $\pm$ 0.001} & 0.796 {\scriptsize $\pm$ 0.007} & 0.941 {\scriptsize $\pm$ 0.001} & 0.976 {\scriptsize $\pm$ 0.001} \\
\midrule
LAMA & 0.913 {\scriptsize $\pm$ 0.001} & 0.881 {\scriptsize $\pm$ 0.010} & 0.962 {\scriptsize $\pm$ 0.003} & 0.984 {\scriptsize $\pm$ 0.001} \\
\bottomrule
\end{tabular}
\end{table}

Table \ref{tab:region6_results} shows the results of the 6-continent classification task. In 6-continent classification, LAMA achieved an accuracy of 0.913, surpassing Self-Reflection (0.847) by 0.066 points. In Macro-F1 as well, LAMA achieved 0.881, with a difference of 0.085 points from Self-Reflection (0.796). Precision@2 reached 0.962 and Precision@3 reached 0.984, demonstrating that LAMA's top prediction candidates have high reliability.

Synthesizing results across the three granularity levels (99 countries, 14 regions, 6 continents), several important trends are observed. First, LAMA achieved the highest performance at all granularities, confirming the generality of the recall-based approach. Second, while overall accuracy improves as granularity becomes coarser, the gap between LAMA and other methods tends to expand. The difference between LAMA and Self-Reflection in 99-country classification was 0.041 points, but it became 0.106 points for 14 regions and 0.066 points for 6 regions. This indicates that LAMA exhibits particularly strong performance in region-level prediction. Third, while the performance gap between neural models and LLM methods tends to shrink as granularity becomes coarser, LAMA consistently maintains a large margin.

\subsection{Ablation Study}

To quantitatively evaluate the contribution of each component of the LAMA framework, we conducted ablation experiments. Table \ref{tab:ablation} shows the performance changes when each component is removed.

\begin{table}[t]
\centering
\caption{Ablation experiment results for LAMA}
\label{tab:ablation}
\begin{tabular}{lccccr}
\toprule
Configuration & Accuracy & Macro-F1 & P@3 & P@5 & $\Delta$Acc \\
\midrule
LAMA (Full) & 0.814 & 0.821 & 0.882 & 0.900 & -- \\
\midrule
w/o Person Agent & 0.812 & 0.818 & 0.878 & 0.897 & $-$0.002 \\
w/o Media Agent & 0.812 & 0.819 & 0.880 & 0.897 & $-$0.002 \\
w/o Completion & 0.815 & 0.822 & 0.850 & 0.857 & $+$0.001 \\
w/o Recall & 0.750 & 0.758 & 0.854 & 0.882 & $-$0.064 \\
\bottomrule
\end{tabular}
\end{table}

The settings for each ablation condition are as follows. w/o Person Agent is a configuration that removes the Person Agent and performs recall using only the Media Agent. Recall of general famous individuals such as politicians, scientists, and business leaders is lost. w/o Media Agent is a configuration that removes the Media Agent and performs recall using only the Person Agent. Recall of athletes and entertainers is lost. w/o Completion is a configuration that removes the completion phase and generates Top-K predictions using only voting from recall results. Only recalled nationalities become prediction candidates, and additional candidate generation by the LLM is not performed. w/o Recall is a configuration that removes both the Person Agent and Media Agent, omitting the entire recall phase. In this configuration, nationality is predicted directly from the name, similar to conventional LLM prompting methods.

From the ablation results, the following important findings were obtained. First, it became clear that the recall mechanism is the most important component for LAMA's performance. When the recall phase was completely removed (w/o Recall), accuracy decreased from 0.814 to 0.750, a drop of 0.064 points. This decrease is overwhelmingly larger compared to when other components are removed, strongly suggesting that LAMA's performance improvement is attributable to indirect reasoning based on recall. The configuration with recall removed is essentially equivalent to conventional LLM direct prediction, and this result empirically demonstrates the effectiveness of the recall approach that explicitly leverages LLM world knowledge.

Second, synergistic effects in the dual-agent configuration were confirmed. When the Person Agent or Media Agent was individually removed, an accuracy decrease of 0.002 points was observed for each. However, when both agents were removed simultaneously (w/o Recall), the decrease was 0.064 points, far exceeding the simple sum of decreases from individual removal (0.004 points). This nonlinear effect indicates that the two agents are not simply contributing independently but are functioning complementarily. Even when only one agent exists, that agent can compensate to some extent for the absence of the other, so performance decrease from individual removal is kept minor. However, when both agents are simultaneously absent, this mutual compensation mechanism is lost, making recall-based prediction completely impossible, resulting in a substantial performance decrease. This result supports the design validity of the dual-agent configuration and indicates that recall comprehensiveness is essential for LAMA's performance.

Third, it became clear that the completion phase contributes to improving Precision@K rather than accuracy. When the completion phase was removed (w/o Completion), accuracy slightly improved to 0.815, while Precision@3 decreased substantially from 0.882 to 0.850, and Precision@5 decreased from 0.900 to 0.857. This result indicates that the completion phase does not affect Top-1 prediction accuracy but plays a role in improving the diversity and quality of candidates from Top-2 onwards. Therefore, adopting the completion phase becomes a design choice dependent on the application scenario. When only Top-1 prediction is needed, the completion phase can be omitted to reduce API call counts, and when presentation of multiple candidates is required, including the completion phase can improve Precision@K.

From these results, it was confirmed that LAMA's performance is primarily supported by the recall mechanism, and the synergistic effects of the dual-agent configuration and the completion phase contribute to improvements in different aspects. In particular, the substantial performance decrease from removing recall empirically supports the core claim of this study—that nationality prediction can be improved by leveraging LLM world knowledge as associative memory.

\subsection{Evaluation on Robustness to Label Frequency}

In nationality prediction tasks, the label occurrence frequency in training data has a significant impact on method performance. High-frequency nationalities are easy to learn due to sufficient sample sizes, while low-frequency nationalities are difficult to predict due to limited samples. This section evaluates the robustness of each method to label frequency.

For evaluation, the 99 nationalities were divided into three bins based on their occurrence frequency in the training data. Head (high frequency) includes the top 33 nationalities, Mid (medium frequency) includes the middle 33 nationalities, and Tail (low frequency) includes the bottom 33 nationalities. For test samples belonging to each bin, accuracy was calculated for each method.

\begin{figure}[t]
\centering
\includegraphics[width=\linewidth]{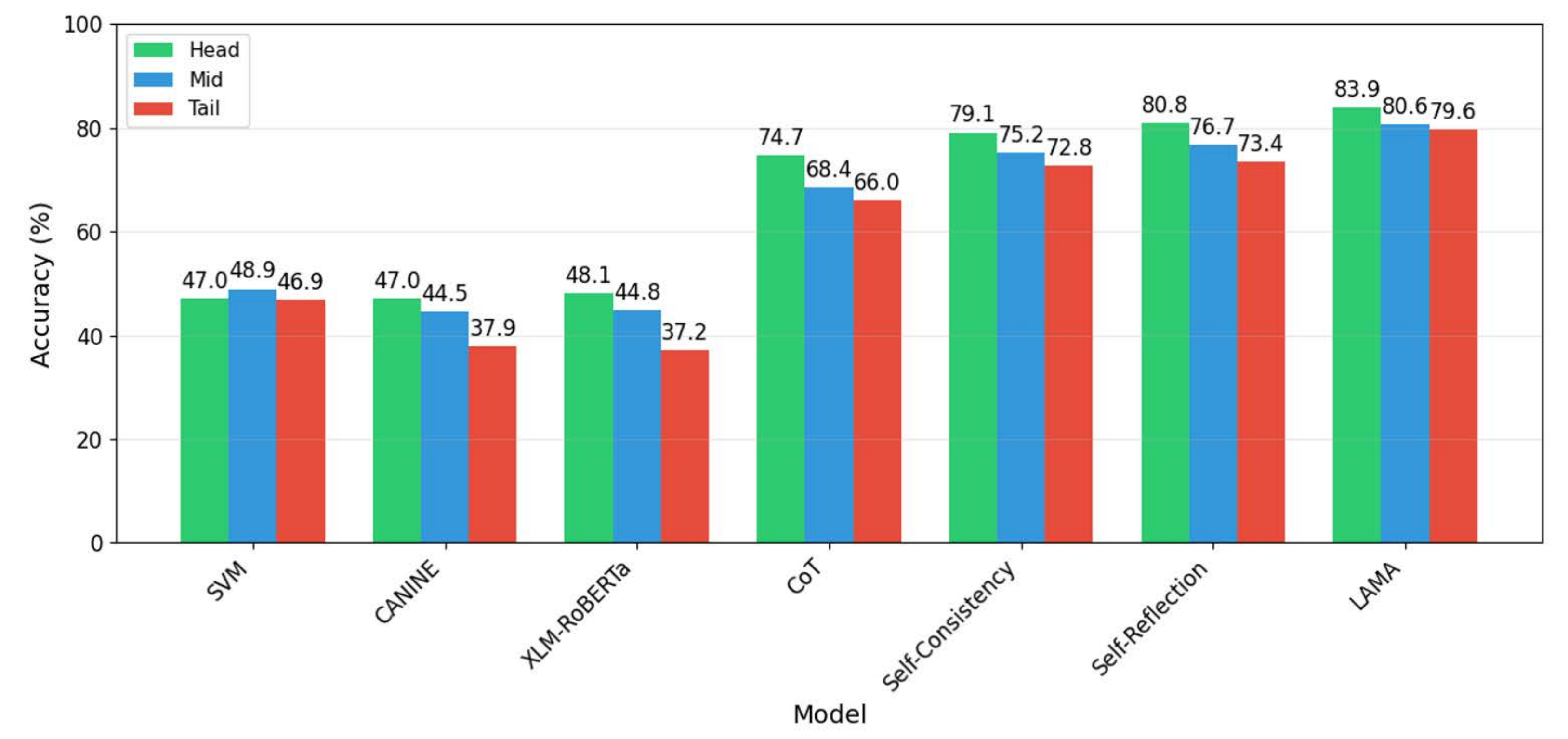}
\caption{Accuracy comparison by frequency bin.}
\label{fig:accuracy_by_bin}
\end{figure}

Figure \ref{fig:accuracy_by_bin} shows the accuracy by frequency bin for each method. CANINE and XLM-RoBERTa show notable performance degradation from Head to Tail. In particular, XLM-RoBERTa shows a decrease of 0.109 points from Head (0.481) to Tail (0.372), with a relative drop rate of 22.6\%. This demonstrates that even pre-trained models have difficulty handling low-frequency labels. On the other hand, LLM methods show overall performance exceeding neural models, but a decreasing trend from Head to Tail is still observed. The relative drop rates for CoT, Self-Consistency, and Self-Reflection were 11.7\%, 8.0\%, and 9.2\%, respectively.

LAMA achieved the highest performance in all bins. With Head (0.839), Mid (0.806), and Tail (0.796), the decrease from Head to Tail was only 0.043 points, with a relative drop rate suppressed to 5.0\%. This is the smallest drop rate among all LLM methods, indicating that LAMA is particularly robust to low-frequency nationalities. Note that SVM shows an exceptionally low relative drop rate (0.3\%), but this is due to uniformly low performance across Head, Mid, and Tail, and does not indicate high robustness. A similar trend was observed for Macro-F1 scores, with LAMA achieving the highest performance in all bins.

LAMA's robustness is considered to be due to characteristics of the recall-based approach. Neural models directly depend on the frequency distribution of training data, so performance degrades when samples of low-frequency nationalities are insufficient. On the other hand, LAMA performs predictions through famous individuals contained in LLM world knowledge, so it is less affected by the frequency distribution of training data. Even for low-frequency nationalities, accurate prediction is possible if prominent individuals of that nationality are contained in the LLM's knowledge. Due to this characteristic, LAMA can mitigate the effects of imbalance in training data and achieve stable performance even for low-frequency nationalities.

This result demonstrates LAMA's practical advantages. In real-world applications, reliable predictions are often required even for low-frequency nationalities. LAMA maintains high performance for high-frequency nationalities while also possessing robustness to low-frequency nationalities, making it a suitable method for a wide range of application scenarios.

\subsection{Error Analysis}

To understand the characteristics of LAMA's prediction errors, we conducted detailed error analysis. This section reveals LAMA's strengths and limitations through analysis of confusion patterns, region-level accuracy, and examination of specific error cases.

Table \ref{tab:confusion_pairs} shows the top 10 confusion pairs for Self-Reflection and LAMA. A confusion pair is a combination of labels where a specific correct label is most frequently incorrectly predicted as another label. The ``Same'' column in the table indicates whether the incorrectly predicted nationality belongs to the same region (14-region classification) as the correct nationality. Additionally, ``Region Match Rate'' represents the proportion of pairs among the top 10 confusion pairs where the predicted nationality and correct nationality belong to the same region. A higher region match rate means that errors are occurring between similar nationalities within the same region, and a lower region match rate means that there are more errors across regions.

\begin{table}[t]
\centering
\caption{Comparison of top 10 confusion pairs. Same indicates whether the predicted nationality belongs to the same region as the correct answer.}
\label{tab:confusion_pairs}
\begin{tabular}{llc|llc}
\toprule
\multicolumn{3}{c|}{Self-Reflection} & \multicolumn{3}{c}{LAMA} \\
True $\rightarrow$ Pred & Count & Same & True $\rightarrow$ Pred & Count & Same \\
\midrule
Tamil $\rightarrow$ Indian & 62 & \checkmark & Tamil $\rightarrow$ Indian & 66 & \checkmark \\
English $\rightarrow$ British & 60 & \checkmark & English $\rightarrow$ British & 53 & \checkmark \\
Welsh $\rightarrow$ British & 32 & \checkmark & Welsh $\rightarrow$ British & 43 & \checkmark \\
Australian $\rightarrow$ American & 27 & & Belarusian $\rightarrow$ Russian & 16 & \checkmark \\
Jamaican $\rightarrow$ American & 23 & & Welsh $\rightarrow$ American & 16 & \\
Belarusian $\rightarrow$ Russian & 22 & \checkmark & Australian $\rightarrow$ American & 16 & \\
Canadian $\rightarrow$ American & 22 & \checkmark & Irish $\rightarrow$ American & 16 & \\
Welsh $\rightarrow$ American & 21 & & Taiwanese $\rightarrow$ Chinese & 15 & \checkmark \\
Cuban $\rightarrow$ Mexican & 21 & & Jamaican $\rightarrow$ American & 14 & \\
Cypriot $\rightarrow$ Greek & 20 & \checkmark & Guinean $\rightarrow$ Australian & 13 & \\
\midrule
\multicolumn{2}{l}{Region Match Rate} & 0.667 & \multicolumn{2}{l}{Region Match Rate} & 0.467 \\
\bottomrule
\end{tabular}
\end{table}

Common confusion patterns for both methods include Tamil→Indian, English→British, and Welsh→British ranking at the top. These are confusions between nationalities with close cultural and historical relationships, representing cases that are inherently difficult to distinguish due to similarity in name patterns. Interestingly, LAMA's region match rate in confusion pairs (0.467) is lower than Self-Reflection's (0.667). At first glance, this appears to indicate that LAMA's performance is inferior. However, in reality, this indicates that as LAMA has reduced ``easy'' errors within the same region, the proportion of ``difficult'' cases that cross regions has relatively increased in the remaining errors.

Table \ref{tab:region_accuracy} shows the breakdown of nationality and region-level accuracy for each method. The region here refers to the 14 regions that classify the 99 nationalities based on geographical proximity. This analysis allows us to understand the proportions of cases where nationality prediction was correct for all samples, cases where nationality was incorrect but region was correct, and cases where both nationality and region were incorrect.

\begin{table}[t]
\centering
\caption{Region-level accuracy analysis}
\label{tab:region_accuracy}
\small
\begin{tabular}{lcccc}
\toprule
Model & Nat Correct & \shortstack{Nat Wrong\\Reg Correct} & \shortstack{Nat Wrong\\Reg Wrong} & \shortstack{Region\\Accuracy} \\
\midrule
SVM & 0.476 & 0.198 & 0.326 & 0.674 \\
CANINE & 0.434 & 0.237 & 0.329 & 0.671 \\
XLM-RoBERTa & 0.437 & 0.231 & 0.332 & 0.668 \\
\midrule
CoT & 0.699 & 0.130 & 0.171 & 0.828 \\
Self-Consistency & 0.758 & 0.103 & 0.139 & 0.861 \\
Self-Reflection & 0.771 & 0.098 & 0.130 & 0.870 \\
\midrule
LAMA & 0.814 & 0.076 & 0.109 & 0.891 \\
\bottomrule
\end{tabular}
\end{table}

Each category in the table means the following. ``Nat Correct'' is the proportion of samples where nationality was correct, ``Nat Wrong, Reg Correct'' is the proportion of samples where nationality was incorrect but the predicted nationality and correct nationality belong to the same region, and ``Nat Wrong, Reg Wrong'' is the proportion of samples where both nationality and region were incorrect. ``Region Accuracy'' is the sum of ``Nat Correct'' and ``Nat Wrong, Reg Correct,'' indicating the proportion of all samples where the prediction belongs to the same region as the correct answer. Note that this Region Accuracy is a metric targeting all samples and differs in calculation scope from the region match rate shown in Table \ref{tab:confusion_pairs} (which targets only top confusion pairs).

LAMA achieved the highest performance in both nationality accuracy (0.814) and Region Accuracy (0.891). Additionally, LAMA's ``Nat Wrong, Reg Wrong'' proportion (0.109) is the lowest, indicating that completely incorrect predictions (both nationality and region incorrect) are fewest.

Qualitative differences are observed between neural models and LLM methods. SVM, CANINE, and XLM-RoBERTa have high ``Nat Wrong, Reg Correct'' proportions of 0.198--0.237. This indicates that neural models have a strong tendency to confuse similar nationalities within the same region (e.g., English and British, Tamil and Indian) without being able to make fine distinctions between nationalities. On the other hand, LLM methods have low proportions of 0.076--0.130, indicating that they can achieve fine classification at the nationality level rather than remaining at coarse region-level classification. LAMA shows this tendency most prominently, with most remaining errors concentrated in difficult cases where both nationality and region differ.

Finally, we analyze the differences in error patterns between Self-Reflection (SR) and LAMA. Table \ref{tab:sr_wrong_lama_correct} shows representative examples where SR was incorrect and LAMA was correct, and Table \ref{tab:sr_correct_lama_wrong} shows representative examples where SR was correct and LAMA was incorrect.

\begin{table}[t]
\centering
\caption{Representative examples where Self-Reflection was incorrect and LAMA was correct. Region $\checkmark$ indicates that SR's prediction belongs to the same region as the correct answer.}
\label{tab:sr_wrong_lama_correct}
\small
\begin{tabular}{lllcl}
\toprule
Name & True & SR Pred & Region $\checkmark$ & LAMA Pred \\
\midrule
Natalie Cook & Australian & American & & Australian \\
Carina Damm & Brazilian & German & & Brazilian \\
Junior Paulo & Samoan & Brazilian & & Samoan \\
Santokh Singh Matharu & Kenyan & Indian & & Kenyan \\
Claudio Nasco & Cuban & Italian & & Cuban \\
Nataliya Moroz & Belarusian & Ukrainian & $\checkmark$ & Belarusian \\
A. A. Khan & Indian & Pakistani & $\checkmark$ & Indian \\
Carmen Brouard & Haitian & Dominican & $\checkmark$ & Haitian \\
\bottomrule
\end{tabular}
\end{table}

\begin{table}[t]
\centering
\caption{Representative examples where Self-Reflection was correct and LAMA was incorrect. Region $\checkmark$ indicates that LAMA's prediction belongs to the same region as the correct answer.}
\label{tab:sr_correct_lama_wrong}
\small
\begin{tabular}{llllc}
\toprule
Name & True (= SR Pred) & LAMA Pred & Region $\checkmark$ \\
\midrule
Jin Chao-chun & Taiwanese & Chinese & $\checkmark$ \\
Marc de Val & Catalan & French & $\checkmark$ \\
Jane Squire & English & British & $\checkmark$ \\
Olo Brown & Samoan & Australian & $\checkmark$ \\
Hussein Ali & Egyptian & Iraqi & $\checkmark$ \\
Guillermo Cerda & Mexican & Chilean & \\
Riad Hammadou & Algerian & French & \\
Mohammed Muftawu & Ghanaian & German & \\
Guy Joseph Bonnet & Haitian & French & \\
\bottomrule
\end{tabular}
\end{table}

In cases where SR was incorrect and LAMA was correct (Table \ref{tab:sr_wrong_lama_correct}), the superiority of the recall-based approach was observed. SR tends to make incorrect inferences across regions based on superficial features of names, such as incorrectly predicting Australian as American and Brazilian as German. In contrast, LAMA reaches accurate predictions by recalling famous individuals with the same name.

On the other hand, in cases where LAMA was incorrect and SR was correct (Table \ref{tab:sr_correct_lama_wrong}), two patterns were observed. First, errors to similar nationalities within the same region (Taiwanese→Chinese, English→British, etc.), which are confusions between nationalities with close cultural and historical relationships. Second, errors pulled by the nationality of recalled famous individuals (Mexican→Chilean, Algerian→French, etc.), which occur when more famous individuals with the same name as the input have different nationalities.

From this error analysis, it became clear that LAMA can reduce errors across regions based on superficial features compared to SR, while having a unique limitation of depending on the nationality distribution of recalled famous individuals. LAMA's errors mainly occur between nationalities with close cultural and historical relationships, concentrated in cases that are inherently difficult to distinguish from name patterns.


\section{Discussion}

\subsection{Key Findings}

The experimental results of this study yielded new insights into attribute prediction tasks from names. First, it became clear that LLMs exhibit higher reliability in concrete example knowledge—``who are famous individuals with this name''—than in abstract knowledge such as inferring nationality from names. Conventional LLM prompting methods (CoT, Self-Reflection, etc.) directly infer nationality from the linguistic features of names. In contrast, LAMA achieved an accuracy improvement of 0.041 points by taking an indirect route through recalling famous individuals. This result suggests that in leveraging LLM knowledge, searching and aggregating concrete examples can be more effective than abstract reasoning in some cases. This finding suggests the effectiveness of approaches that go through related examples rather than direct question answering for knowledge extraction tasks using LLMs in general, providing new guidelines for prompt design.

Second, it was confirmed that recall-based approaches have the characteristic of not depending on training data distributions. Neural models showed relative drops of 19.5--22.6\% between Head and Tail, whereas LAMA was suppressed to 5.0\%. This means that because LAMA depends on LLM world knowledge as an external knowledge source, it is less affected by training data imbalance. This characteristic is particularly important for tasks involving low-resource languages or minority classes, indicating the possibility of achieving high performance by leveraging LLM world knowledge even in situations where collecting training data is difficult.

Third, it was demonstrated that combining multiple agents specialized in different knowledge domains produces effects greater than simple addition. The decrease when individually removing the Person Agent or Media Agent was 0.002 points each, but removing both resulted in a decrease of 0.064 points. This nonlinear synergistic effect suggests the existence of a mechanism where one agent compensates for recall failures of the other. This finding supports the effectiveness of combining multiple specialized agents rather than a single general-purpose agent in multi-agent system design. It can be utilized as a design guideline for agent partitioning based on knowledge domains in future LLM agent research.

Fourth, it became clear that optimal strategies differ between Top-1 prediction and Top-K prediction. Removing the completion phase slightly improved Top-1 accuracy, but Precision@5 decreased by 0.043 points. This result indicates that aggregation of recall results alone is sufficient for Top-1 prediction, and additional completion processing is effective for Top-K prediction. This finding means that system configuration can be optimized according to application scenarios, enabling flexible operation where API calls are reduced when only Top-1 prediction is needed, and the completion phase is added when multiple candidates are required.

Synthesizing these findings, a new paradigm for leveraging LLM world knowledge emerges. Conventional LLM utilization has predominantly employed the approach of having models directly answer questions, but this study demonstrated the effectiveness of an indirect approach that has models recall related concrete examples and aggregates that information. This multi-agent system is considered applicable not only to nationality prediction but to any attribute prediction task where concrete examples are contained in LLM knowledge (industry prediction from organization names, country of origin prediction from product names, climate zone prediction from place names, etc.).

\subsection{Limitations}

This study has several limitations. First, this study conducted evaluation using only GPT-4.1-mini, and generalizability to other LLMs has not been verified. Different LLMs (Claude, Gemini, Llama, etc.) may have different scopes of world knowledge and recall capabilities, and the extent to which LAMA's performance depends on the model is unknown. In particular, open-source models and smaller models may have limited knowledge about famous individuals, raising concerns that the effectiveness of recall-based approaches may decrease. Future research requires comparative evaluation using multiple LLMs.

Second, LAMA's performance depends on the distribution of famous individuals contained in LLM world knowledge. LLM training data tends to be biased toward English-speaking and Western countries, and knowledge about famous individuals from specific nationalities or regions may be insufficient. In error analysis, error patterns where LAMA was pulled by the nationality of recalled famous individuals were observed, which may be attributable to knowledge bias. This limitation is related to the fundamental problem of bias in LLM training data, and while not unique to LAMA, it is an important factor that should be considered in practical deployment.

Third, this study targeted only nationality prediction tasks, and applicability to other attribute prediction tasks has not been experimentally verified. As stated in Key Findings, recall-based approaches are considered applicable to other tasks, but effectiveness may differ depending on task characteristics (attribute ambiguity, existence of related famous individuals, knowledge coverage, etc.). Future research requires evaluation of LAMA on diverse attribute prediction tasks.


\section{Conclusion}

In this study, we proposed LAMA, a novel attribute prediction framework that leverages LLM world knowledge as associative memory. Rather than directly inferring nationality from names, LAMA effectively elicits LLMs' concrete factual knowledge through indirect reasoning that recalls famous individuals with the same name and aggregates their nationalities. Through the combination of a dual-agent architecture specialized in different knowledge domains, voting based on recall results, and conditional completion processing, we achieved high-accuracy and robust nationality prediction.

On a nationality prediction task spanning 99 countries, LAMA achieved an accuracy of 0.817, substantially outperforming both conventional LLM prompting methods and neural models. Through our experiments, the following important findings were obtained. First, LLMs exhibited higher reliability in recalling concrete famous individuals than in applying abstract linguistic rules, demonstrating that indirect reasoning is more effective than direct reasoning. Second, recall-based approaches are robust to low-frequency nationalities because they do not depend on training data frequency distributions, with a relative drop rate between Head and Tail of 5.0\%, the smallest among all methods. Third, the dual-agent architecture showed synergistic effects far exceeding individual removal, confirming the complementary functioning of multiple agents specialized in different knowledge domains.

Future research needs to verify LAMA's generalizability through evaluation using multiple LLMs (Claude, Gemini, open-source models, etc.). Additionally, it is important to explore the generality of the indirect reasoning paradigm based on associative memory through application to attribute prediction tasks other than nationality prediction (industry prediction from organization names, country of origin prediction from product names, etc.). Furthermore, to mitigate the error pattern unique to LAMA that depends on the nationality distribution of recalled famous individuals, investigation of reliability evaluation of recall results and aggregation methods through multiple recall attempts is required.

\backmatter

\bmhead{Supplementary information}
Not applicable.

\bmhead{Acknowledgements}
The authors would like to thank The Nippon Foundation HUMAI Program for supporting this study and for providing a research environment that enabled the completion of this work. 

\section*{Declarations}

\begin{itemize}
\item \textbf{Funding:} This work was supported by The Nippon Foundation HUMAI Program.
\item \textbf{Conflict of interest:} The author has no conflicts of interest to declare.
\item \textbf{Ethics approval:} Not applicable.
\item \textbf{Consent to participate:} Not applicable.
\item \textbf{Consent for publication:} Not applicable.
\item \textbf{Data availability:} The name2nat dataset used in this study is publicly available at \url{https://github.com/Kyubyong/name2nat}.
\item \textbf{Code availability:} Code will be made available upon reasonable request.
\item \textbf{Author contribution:} The author conducted all experiments, performed the analysis, and wrote the manuscript.
\end{itemize}









\bibliography{sn-bibliography}

@inproceedings{1,
  author    = {Manvi, Rohin and Khanna, Samar and Mai, Gengchen and Burke, Marshall and Lobell, David and Ermon, Stefano},
  title     = {{GeoLLM}: Extracting Geospatial Knowledge from Large Language Models},
  booktitle = {Proceedings of the International Conference on Learning Representations},
  year      = {2024},
  doi       = {10.48550/arXiv.2310.06213}
}

@inproceedings{2,
  author    = {Wei, J. and Wang, X. and Schuurmans, D. and Bosma, M. and Ichter, B. and Xia, F.},
  title     = {Chain-of-thought prompting elicits reasoning in large language models},
  booktitle = {Proceedings of the International Conference on Neural Information Processing Systems},
  pages     = {24824--24837},
  year      = {2022},
  doi       = {10.48550/arXiv.2201.11903}
}

@inproceedings{3,
  author    = {Shinn, N. and Cassano, F. and Berman, E. and Gopinath, A. and Narasimhan, K. and Yao, S.},
  title     = {Reflexion: Language agents with verbal reinforcement learning},
  booktitle = {Proceedings of the International Conference on Neural Information Processing Systems},
  pages     = {8634--8656},
  year      = {2023},
  doi       = {10.48550/arXiv.2303.11366}
}

@article{4,
  author    = {Webber, R.},
  title     = {Researching behavioural differences among ethnic minority groups: The case for inferring ethnicity on the basis of people's names},
  journal   = {International Journal of Market Research},
  volume    = {52},
  number    = {2},
  pages     = {191--215},
  year      = {2010},
  doi       = {10.2501/s147078530920117x}
}

@article{5,
  author    = {Kandt, J. and Longley, P. A.},
  title     = {Ethnicity estimation using family naming practices},
  journal   = {PloS One},
  volume    = {13},
  number    = {8},
  pages     = {e0201774},
  year      = {2018},
  doi       = {10.1371/journal.pone.0201774}
}

@inproceedings{6,
  author    = {Treeratpituk, P. and Giles, C. L.},
  title     = {Name-ethnicity classification and ethnicity-sensitive name matching},
  booktitle = {Proceedings of the AAAI Conference on Artificial Intelligence},
  volume    = {26},
  number    = {1},
  pages     = {1141--1147},
  year      = {2012},
  doi       = {10.1609/aaai.v26i1.8324}
}

@article{7,
  author    = {Xie, F.},
  title     = {rethnicity: An {R} package for predicting ethnicity from names},
  journal   = {SoftwareX},
  volume    = {17},
  pages     = {100965},
  year      = {2022},
  doi       = {10.1016/j.softx.2021.100965}
}

@article{8,
  author    = {Clark, J. H. and Garrette, D. and Turc, I. and Wieting, J.},
  title     = {{CANINE}: Pre-training an efficient tokenization-free encoder for language representation},
  journal   = {Transactions of the Association for Computational Linguistics},
  volume    = {10},
  pages     = {73--91},
  year      = {2022},
  doi       = {10.1162/tacl_a_00448}
}

@inproceedings{9,
  author    = {Conneau, A. and Khandelwal, K. and Goyal, N. and Chaudhary, V. and Wenzek, G. and Guzm{\'a}n, F.},
  title     = {Unsupervised cross-lingual representation learning at scale},
  booktitle = {Proceedings of the Annual Meeting of the Association for Computational Linguistics},
  pages     = {8440--8451},
  year      = {2020},
  address   = {Online},
  publisher = {Association for Computational Linguistics},
  doi       = {10.18653/v1/2020.acl-main.747}
}

@article{10,
  author  = {Inoshita, K},
  title   = {Nationality and region prediction from names: A comparative study of neural models and large language models},
  journal = {arXiv preprint arXiv:2601.08692},
  year    = {2026},
  doi     = {10.48550/arXiv.2601.08692}
}

@article{11,
  author    = {Mateos, P.},
  title     = {A review of name-based ethnicity classification methods and their potential in population studies},
  journal   = {Population, Space and Place},
  volume    = {13},
  number    = {4},
  pages     = {243--263},
  year      = {2007},
  doi       = {10.1002/psp.457}
}

@article{12,
  author    = {Voicu, I.},
  title     = {Using first name information to improve race and ethnicity classification},
  journal   = {Statistics and Public Policy},
  volume    = {5},
  number    = {1},
  pages     = {1--13},
  year      = {2018},
  doi       = {10.1080/2330443x.2018.1427012}
}

@inproceedings{13,
  author    = {Ye, J. and Han, S. and Hu, Y. and Coskun, B. and Liu, M. and Qin, H.},
  title     = {Nationality classification using name embeddings},
  booktitle = {Proceedings of the ACM Conference on Information and Knowledge Management},
  year      = {2017},
  address   = {Singapore, Singapore},
  publisher = {ACM},
  doi       = {10.1145/3132847.3133008}
}

@inproceedings{14,
  author    = {Lee, J. and Kim, H. and Ko, M. and Choi, D. and Choi, J. and Kang, J.},
  title     = {Name Nationality Classification with Recurrent Neural Networks},
  booktitle = {Proceedings of the International Joint Conference on Artificial Intelligence},
  pages     = {2081--2087},
  year      = {2017},
  doi       = {10.24963/ijcai.2017/289}
}

@article{15,
  author    = {Hur, Y.},
  title     = {Malaysian Name-based Ethnicity Classification using {LSTM}},
  journal   = {KSII Transactions on Internet and Information Systems},
  volume    = {16},
  number    = {12},
  pages     = {3855--3867},
  year      = {2022},
  doi       = {10.3837/tiis.2022.12.004}
}

@inproceedings{16,
  author    = {An, H. and Acquaye, C. and Wang, C. and Li, Z. and Rudinger, R.},
  title     = {Do large language models discriminate in hiring decisions on the basis of race, ethnicity, and gender?},
  booktitle = {Proceedings of the Annual Meeting of the Association for Computational Linguistics},
  pages     = {386--397},
  year      = {2024},
  address   = {Bangkok, Thailand},
  publisher = {Association for Computational Linguistics},
  doi       = {10.18653/v1/2024.acl-short.37}
}

@inproceedings{17,
  author    = {Sakunkoo, A. and Sakunkoo, J.},
  title     = {Name of thrones: How do {LLMs} rank student names in status hierarchies based on race and gender?},
  booktitle = {Proceedings of the Workshop on Innovative Use of NLP for Building Educational Applications},
  pages     = {697--707},
  year      = {2025},
  address   = {Vienna, Austria},
  publisher = {Association for Computational Linguistics},
  doi       = {10.18653/v1/2025.bea-1.50}
}

@article{18,
  author  = {Phonchai, T and Siripong, S and Patterson, N and Campbell, O},
  title   = {Large Language Models for Zero-shot Multicultural Name Recognition},
  journal = {arXiv preprint arXiv:2507.04149},
  year    = {2025},
  doi     = {10.48550/arXiv.2507.04149}
}

@inproceedings{19,
  author    = {Wang, X. and Wei, J. and Schuurmans, D. and Le, Q. and Chi, E. and Narang, S.},
  title     = {Self-consistency improves chain of thought reasoning in language models},
  booktitle = {Proceedings of the International Conference on Learning Representations},
  year      = {2023},
  doi       = {10.48550/arXiv.2203.11171}
}

@article{20,
  author  = {Chen, P and Chen, S and Wang, M and Leong, S and Fung, P and Bernales, V},
  title   = {Schema for In-Context Learning},
  journal = {arXiv preprint arXiv:2510.13905},
  year    = {2025},
  doi     = {10.48550/arXiv.2510.13905}
}

@inproceedings{21,
  author  = {Budagam, D and Kumar, A and Khoshnoodi, M and Kj, S and Jain, V and Chadha, A},
  title   = {Hierarchical Prompting Taxonomy: A universal evaluation framework for large language models aligned with human cognitive principles},
  booktitle = {Proceedings of the KDD 2025 Workshop on Prompt Optimization},
  address   = {Toront, Canada},
  year    = {2025},
  doi     = {10.48550/arXiv.2406.12644}
}

@article{22,
  author  = {Sumers, T and Yao, S and Narasimhan, K and Griffiths, T},
  title   = {Cognitive Architectures for Language Agents},
  journal = {Transactions on Machine Learning Research},
  year    = {2023},
  doi     = {10.48550/arXiv.2309.02427}
}

@article{23,
  author  = {Kirk, J and Wray, R and Laird, J},
  title   = {Exploiting language models as a source of knowledge for cognitive agents},
  journal = {Proceedings of the AAAI Fall Symposium Series},
  pages     = {286--294},
  volume    = {2},
  number    = {1},
  year    = {2023},
  doi     = {10.48550/arXiv.2310.06846}
}

@article{24,
  author  = {Shan, L and Luo, S and Zhu, Z and Yuan, Y and Wu, Y},
  title   = {Cognitive memory in large language models},
  journal = {arXiv preprint arXiv:2504.02441},
  year    = {2025},
  doi     = {10.48550/arXiv.2504.02441}
}

@article{25,
  author  = {Inoshita, K and Mizuno, S},
  title   = {World model inspired sarcasm reasoning with large language model agents},
  journal = {arXiv preprint arXiv:2512.24329},
  year    = {2025},
  doi     = {10.48550/arXiv.2512.24329}
}

@misc{26,
  author       = {Park, Kyubyong},
  title        = {name2nat: a {Python} package for nationality prediction from a name},
  year         = {2020},
  howpublished = {GitHub repository},
  url          = {https://github.com/Kyubyong/name2nat}
}

\end{document}